\documentclass[twocolumn,twocolappendix]{aastex701}
\usepackage{array}
\usepackage{xcolor}
\usepackage{subcaption}
\usepackage{booktabs}
\usepackage{tabularx,colortbl}
\usepackage{amsmath}
\usepackage{tikz}
\usepackage{float}
\usepackage{xspace}

\newcommand{\authz}[1]{#1}

\newcommand{\rubr}{$RuBR$\xspace}
\newcommand{\rubrc}{$RuBR_{comb}$\xspace}
\newcommand{\rubrl}{$RuBR_{loc}$\xspace}
\newcommand{\rubrd}{$RuBR_{DA}$\xspace}

\usetikzlibrary{shapes.geometric, arrows.meta, positioning}

\tikzset{
    box/.style={
        rectangle,
        draw=orange!70!black,
        fill=orange!5,
        thick,
        text width=4.5cm,
        minimum height=2cm,
        align=center,
        font=\small
    },
    arrow/.style={
        -Stealth,
        thick,
        orange!70!black,
        line width=0.8pt
    }
}

\newcommand{\feats}{$\mathcal{F}^{(i)}$}

\begin{document}

\title{Identifying Gems from Roman RAPIDly}
\correspondingauthor{Ashish~A.~Mahabal}

\author[0009-0007-9530-5577]{Karan Gandhi}
\affiliation{Department of Computer Science and Engineering, Indian Institute of Technology, Gandhinagar, India}
\email{23110157@iitgn.ac.in}

\author[0000-0003-2242-0244]{Ashish~A.~Mahabal}
\affiliation{Division of Physics, Mathematics, and Astronomy, California Institute of Technology, Pasadena, CA 91125, USA}
\affiliation{Center for Data Driven Discovery, California Institute of Technology, Pasadena, CA 91125, USA}
\email[show]{aam@astro.caltech.edu}

\author[0000-0001-5754-4007]{Jacob E. Jencson}
\affiliation{IPAC, California Institute of Technology, 1200 E. California Blvd, Pasadena, CA 91125, USA}
\email{jjencson@ipac.caltech.edu}

\author[0000-0003-2451-5482]{Russ R. Laher}
\affiliation{IPAC, California Institute of Technology, 1200 E. California Blvd, Pasadena, CA 91125, USA}
\email{laher@ipac.caltech.edu}

\author[0000-0001-7648-4142]{Ben Rusholme}
\affiliation{IPAC, California Institute of Technology, 1200 E. California Blvd, Pasadena, CA 91125, USA}
\email{rusholme@ipac.caltech.edu}

\author[0000-0003-1710-9339]{Lin Yan}
\affiliation{Caltech Optical Observatories, California Institute of Technology, Pasadena, CA 91125, USA}
\email{lyan@caltech.edu}

\author[0000-0003-0778-0321]{Ryan M. Lau}
\affiliation{IPAC, California Institute of Technology, 1200 E. California Blvd, Pasadena, CA 91125, USA}
\email{ryanlau@ipac.caltech.edu}

\author[0000-0001-9038-9950]{Schuyler D. Van Dyk}
\affiliation{IPAC, California Institute of Technology, 1200 E. California Blvd, Pasadena, CA 91125, USA}
\email{vandyk@ipac.caltech.edu}


\author[0000-0002-5619-4938]{Mansi M. Kasliwal}
\affiliation{Division of Physics, Mathematics, and Astronomy, California Institute of Technology, Pasadena, CA 91125, USA}
\email{mansi@astro.caltech.edu}

\begin{abstract}

The Nancy Grace Roman Space Telescope (Roman), set for launch as early as September 2026, will conduct wide-field infrared imaging surveys with unprecedented spatial resolution and cadence, enabling the discovery of millions of astronomical transients. Hence, it is necessary to have automated pipelines for generating alerts in place so that the telescope can begin discovering reliable transients and variable objects soon after it is launched. However, no real Roman data currently exist, making the development of such pipelines difficult. 
In this work, we present \authz{a machine learning model \rubr} and a general methodology for distinguishing genuine transient and variable detections from spurious (bogus) detections within the RAPID pipeline. 
\authz{In particular, we present three models using this methodology: \rubrc trained and tested on combined locally injected and OpenUniverse2024 transients, \rubrl trained on locally injected transients and tested on OpenUniverse2024 transients, and \rubrd that combines locally injected transients with a fraction of OpenUniverse2024 transients in domain-adaptation mode for training.} \authz{This paves the way for} strategies to adapt the \authz{\rubrc} model to real observations in the absence of any ground-truth labels during the early phases of the Roman mission.
While the image differencing pipeline continues to be improved, our experimental results demonstrate the effectiveness of the proposed approach and its promise for robust real-bogus classification in the Roman era. 
\end{abstract}


\section{Introduction and Context} \label{sec:Intro}

The upcoming Nancy Grace Roman Space Telescope (Roman, hereafter) will obtain pointed observations as well as multiple multi-epoch surveys, leading to the detection of millions of astronomical transients, enabling science in uncharted parameter spaces. Roman is NASA’s next-generation infrared space telescope, currently in development and scheduled to launch into a Sun-Earth L2 orbit as early as September 2026. Equipped with the Wide-Field Instrument (WFI), Roman will deliver images with sharpness comparable to those of the Hubble Space Telescope but over a much larger field of view, 0.28 square degrees, 100 times larger than Hubble's imaging cameras \citep{collaboration2025openuniverse2024}.

Roman’s High Latitude Time Domain Survey (HLTDS), Galactic Bulge Time Domain Survey (GBTDS), and High Latitude Wide Area Survey (HLWAS) will capture a vast number of transient and variable events, including supernovae, tidal disruption events, kilonovae, etc. This data will contribute to advancements in astrophysics, like detecting and studying high-redshifted Tidal Disruption Events (TDEs) \citep{karmen2023roman}, improving our understanding and measurement of supernovae, leading to more precise cosmological constraints \citep{hounsell2023measuring}, etc. To fully harness the potential of this data, efficient methods for identifying and classifying these transients and variables must be developed. A common real–bogus classification framework can be applied across these surveys; however, the extremely high source density and crowding in GBTDS are expected to make it the most challenging environment for transient discrimination. Developing reliable transient pipelines in advance is therefore critical to ensure that Roman can begin discovering and classifying transients as soon as repeat observations become available at the start of operations. 

A large number of surveys detect transients in the image domain using image differencing \citep{zackay2016proper,alard1998method} and then running source detection algorithms \citep{bertin1996sextractor,stetson1987daophot}. This process isolates variable or new sources, such as supernovae or tidal disruption events, as residuals in the resulting difference image. However, subtraction imperfections, detector artifacts, and cosmic rays can generate a large number of spurious detections. Hence, in many surveys \citep[][etc]{mahabal2019machine, duev2019real, cabrera2017deep}, machine learning–based real–bogus classifiers have proven essential in filtering out false positives from the enormous stream of candidate detections produced by image differencing pipelines. 

However, unlike existing surveys that have access to real data for training their models, Roman has not yet been launched, and therefore, no such data currently exists. Hence, using simulated data, we present in this paper a methodology for distinguishing spurious (bogus) detections from genuine candidates within the alert stream, even in highly imbalanced cases where the number of bogus detections greatly exceeds the number of true ones. We also propose a strategy to adapt the models, once real data becomes available, without requiring ground truth labels. This is necessary because real Roman images are expected to exhibit artifacts that differ from those in our simulated data. This enables astronomers wanting to use these alerts for follow-up observations to make more informed and reliable decisions as soon as the telescope is launched.

We begin by providing an overview of the RAPID\footnote{\authz{RAPID stands for Roman Alerts Promptly from Image Differencing and is one of the Program Infrastructure teams (PIT) for Roman.}} pipeline in Section~\ref{sec:pipe-overview}. Section~\ref{sec:dataset} describes the dataset used for training the models. In Section~\ref{sec:arch} and Section~\ref{sec:training}, we give a detailed description of the proposed model architecture and training procedure. We evaluate our model’s performance against several widely used baseline models in Section~\ref{sec:exp-results}, discuss strategies for adapting the models to real-world data and the associated experiments in Section~\ref{sec:domain-adv}. Finally, we present ablation studies in Section~\ref{sec:ablation},  and we conclude with a summary and directions for future work in Section~\ref{sec:future-work}.

\section{Pipeline Overview}\label{sec:pipe-overview}

\begin{figure*}
    \centering
    \begin{tikzpicture}[node distance=1cm and 0.8cm]
        \node[box] (science) {Science and Reference Images};
        \node[box, right=of science] (cutouts) {Cutouts are made around the detections};
        \node[box, right=of cutouts] (quads) {Cutout Quads are fed into the Real Bogus Model to filter out the Bogus detections};
        
        \node[box, below=of science] (projected) {Projected and subtracted using ZOGY/SFFT};
        \node[box, below=of cutouts] (diff) {Diff/SCORR Image is passed through a detector like SExtractor or PSF Photutils};
        \node[box, below=of quads] (final) {Final Alerts};
        
        \draw[arrow] (science) -- (projected);
        \draw[arrow] (projected) -- (diff);
        \draw[arrow] (diff) -- (cutouts);
        \draw[arrow] (cutouts) -- (quads);
        \draw[arrow] (quads) -- (final);
        
    \end{tikzpicture}
    \caption{\authz{Overview of the RAPID alert-generation pipeline. Science and reference images are first aligned and subtracted using ZOGY/SFFT. Source detection is then performed on the difference image using SExtractor or PSF-based Photutils. Cutouts of size $64\times64$ centered on the detections are passed to the Real–Bogus model to reject spurious candidates, yielding the final alert stream.}}
    \label{fig:pipeline}
\end{figure*}
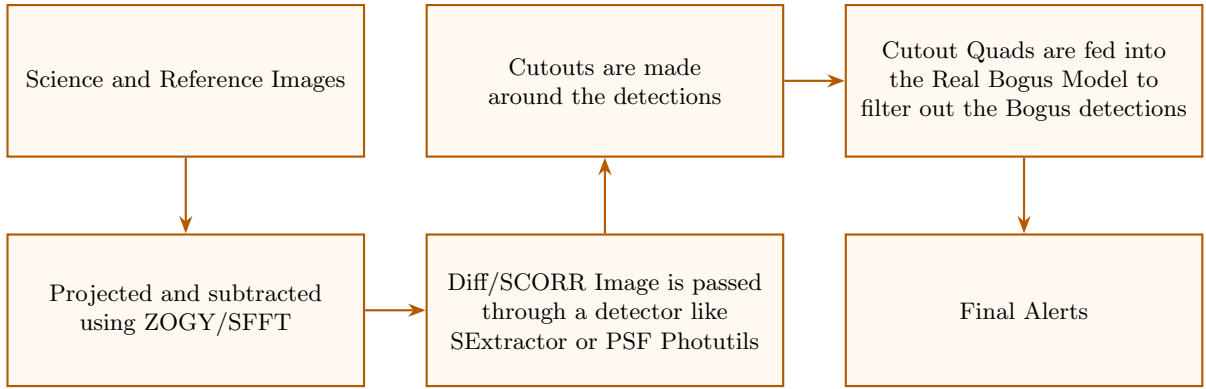

Level-2 (L2) data products are calibrated, two-dimensional rate images expressed in instrumental units of DN/s. These are generated from Level-1 (L1) ramp data by the Exposure Pipeline in romancal \citep{2022rdox.rept......}. Once the level-2 processed images are made available by Roman, 
RAPID
will run the differencing pipeline\footnote{https://caltech-ipac-rapid.readthedocs.io/en/latest/index.html} to create and difference the reference and science images using image differencing algorithms like ZOGY \citep{zackay2016proper}, or SFFT \citep{hu2022image}. Once these difference images have been generated, we run a source extractor on the difference image to generate candidate alerts (see Figure~\ref{fig:pipeline}). The candidate alerts can have a large number of bogus events, especially during the early days of real observations. 

A confidence will be provided so that the threshold can be adjusted by the end users for themselves. Once the bogus events are filtered out, there can be various downstream models, e.g., light curves of these alerts can be extracted and fed into another model, for further classification into many leaf nodes, \authz{for} variable and transient subclassification. Our models will first be trained on simulated data and then adapted to the Roman data using domain adversarial training. Once we have ground truth labels for some of the objects that we identified, we can train the models on the Roman data to greatly improve results.

\section{Dataset}\label{sec:dataset}

\begin{figure*}
    \centering
    \begin{subfigure}{1\linewidth}
        \centering
        \includegraphics[width=\linewidth]{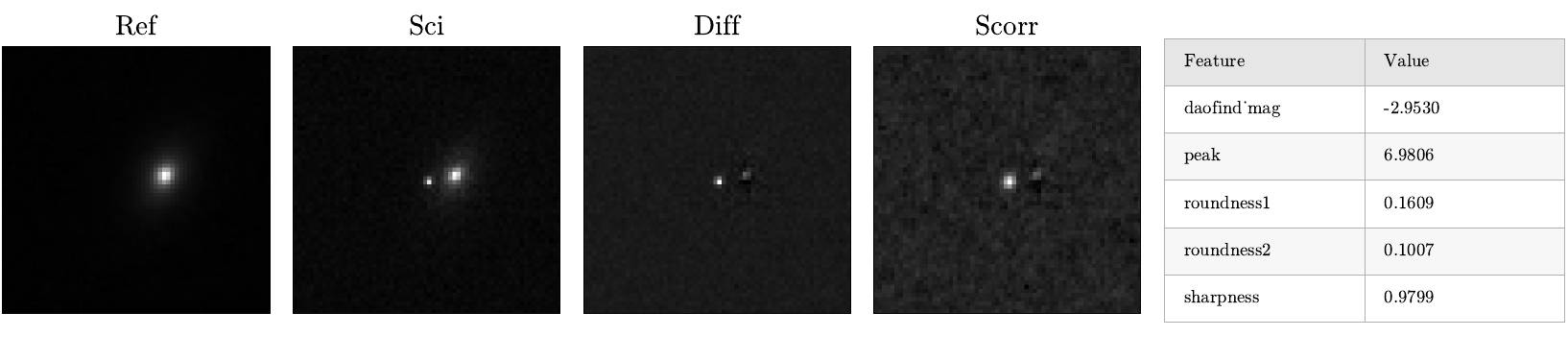}
        \caption{True Positive Example.}
        \label{fig:detections-a}
    \end{subfigure}
    \begin{subfigure}{1\linewidth}
        \centering
        \includegraphics[width=\linewidth]{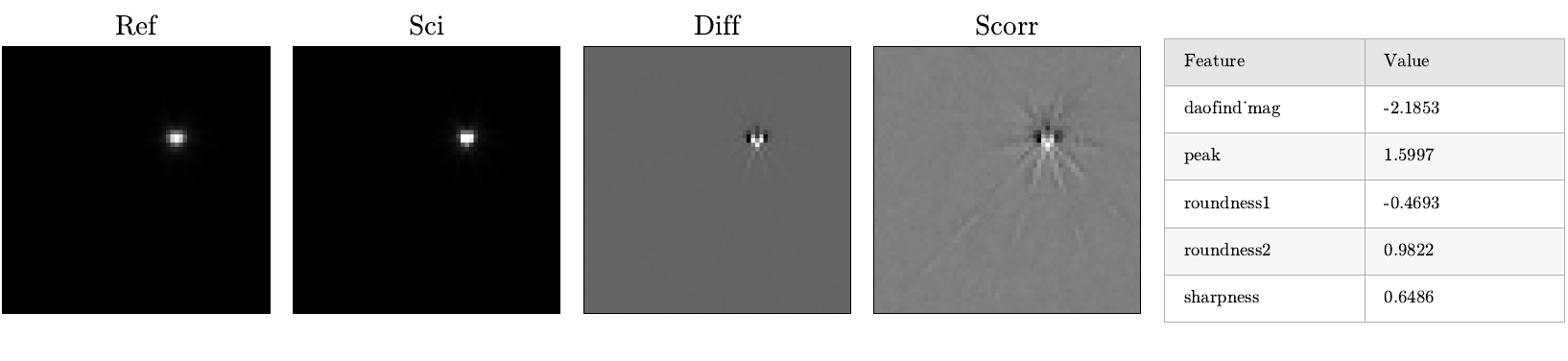}
        \caption{False Positive Example.}
        \label{fig:detections-b}
    \end{subfigure}
    
    \caption{\authz{Example Science, Reference, Difference, and SCORR image quadruplets from the dataset along with the features associated with the objects. Panels (a) and (b) show, respectively, a true and a false transient detection produced by the DAOFind algorithm.}}
    \label{fig:detections-eg}
\end{figure*}

\begin{figure*}
\gridline{
  \fig{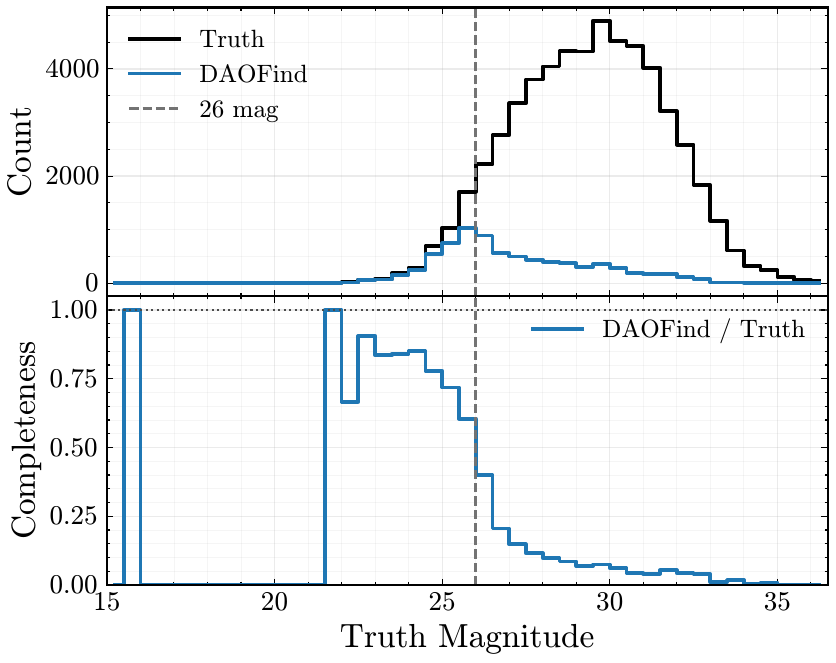}{0.49\textwidth}{(a) DAOFind detections vs.\ AB Magnitude.}
  \fig{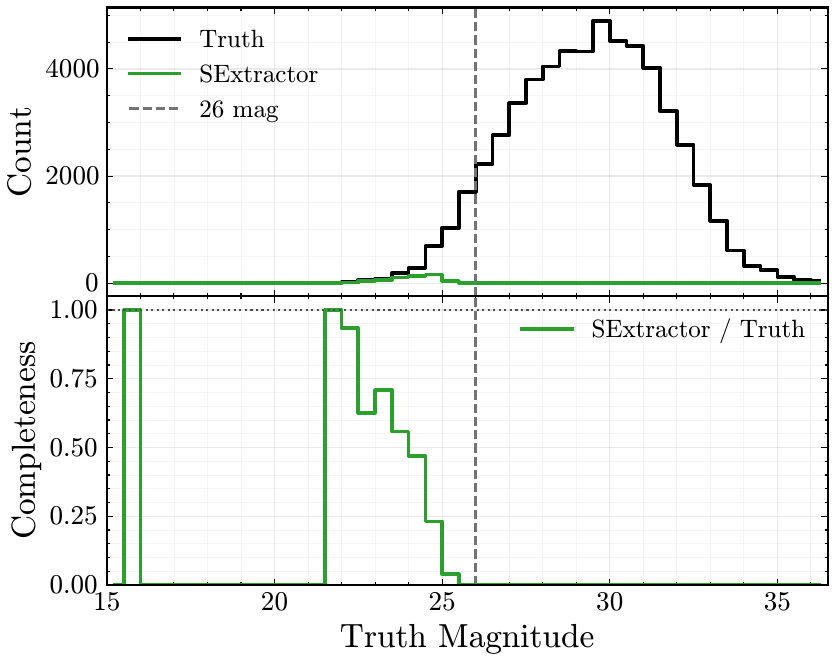}{0.49\textwidth}{(b) SExtractor detections vs.\ AB magnitude.}
}
    \caption{\authz{Comparison of detections as a function of AB magnitude for cross-matched sources using (a) the DAOFind algorithm and (b) SExtractor on a subset of the OpenUniverse2024 simulations with default parameters. SExtractor is more conservative, yielding fewer detections but a higher purity (real-to-bogus ratio). DAOFind, while more complete, suffers from a substantially higher false-positive rate, particularly beyond $\sim$26 mag, where detections are often associated with noise fluctuations near transients that are too faint to be reliably detected. Because of the smaller recall using SExtractor, and because that part of the pipeline was still being improved, we concentrate hereafter on the DAOFind detections.}}
    \label{fig:detections-vs-mag}
\end{figure*}

Since Roman has not been launched yet, we've been working with simulated data from the OpenUniverse2024 (OU24 hereafter) simulations \citep{collaboration2025openuniverse2024}. The OU24 dataset provides simulated imaging data for approximately 70 deg$^2$,  \authz{including implementations of reference survey designs for planned Roman surveys.} 
In this work, we focus on the Roman survey portion of the OU24 simulations, a smaller area coverage ($\approx$12 deg$^2$). \authz{In particular, we use the simulated reference survey implementation of the HLTDS\footnote{For current information on the planned implementation of the actual HLTDS, see https://roman-docs.stsci.edu/roman-community-defined-surveys/high-latitude-time-domain-survey}. The two other Roman Core Community Surveys, viz. HLWAS and GBTDS will have similar processes applied for image subtraction and real-bogus determination, though their cadence and object densities will differ.} The OU24 comprises 1.39 million simulated transient events with 312 million spectral energy distributions (SEDs) spanning a Modified Julian Date (MJD) range from $61444$ to $63269$. Additionally, it incorporates updated optical and near-infrared transient models tailored to Roman’s observational capabilities. The dataset has a few types of explosive events, tidal disruption events, etc., but no simulated variable stars or unassociated transients. 

The simulations also include sample bright \authz{transient} sources near bright galaxies having a random absolute magnitude between $-22$ and \authz{$-17$} and a random redshift between $0$ and $2$. \authz{These events were injected into a ``post-survey'' set of images, $>$280 days after the last MJD for the other classes of astropysical transients.} These bright sources can be used for more detailed studies on image differencing and can help in improving our models. \authz{
This set of post-survey images was used to build references for RAPID pipeline testing; hence, these transients appear as negative sources in our differences (a situation that also occurs when sources fade). Our training models are aware of such sources, but in this work, we have not separately tested for them.}

 We also added new injections locally on top of these simulated images, with $100$ injections per image, irrespective of filters. The injected sources have magnitudes ranging from $21$ to \authz{$28$} and are associated with galaxies. This was done to test how our models perform on data containing artifacts different from those in the training set, since the real data are very likely to include artifacts that differ from our original simulations (as discussed later in Section \ref{sec:domain-adv}).

To make our training dataset, we first create reference images for each science image and difference these images. Once these images are differenced, we pass them through a source detection algorithm like SExtractor \citep{bertin1996sextractor} or the DAOFind algorithm \citep{stetson1987daophot}. We used the photutils package for the DAOFind algorithm. Once we have these detections, we cross-match them with the transients present in the truth files, and if there is a transient present in the source file within a radius of $k$ pixels (in this case $k = 2$), we mark that detection as a true detection, and if no transient is present in the radius, we mark it as a false detection. Now, once we have these true positive examples and false positive examples, we make $L\times L$ (in this case we chose $L= 64$ pixels) sized cutouts of the (i) Science, (ii) Reference, (iii) Difference, and (iv) SCORR Image (the SCORR image is essentially a significance or score map used to highlight potential transient objects relative to the local noise characteristics \citep{zackay2016proper} centred on these detections. We bundle them to form one training example of shape $x^{(i)} \in \mathbb{R} ^{L\times L\times 4}$, having a label $y^{(i)} = 1$ if the example if of a correctly detected transient and $y^{(i)} = 0$ if it is a false example of a transient. This dataset is then split into training, validation, and testing sets.

\begin{figure*}
    \centering
    \includegraphics[width=.9\linewidth]{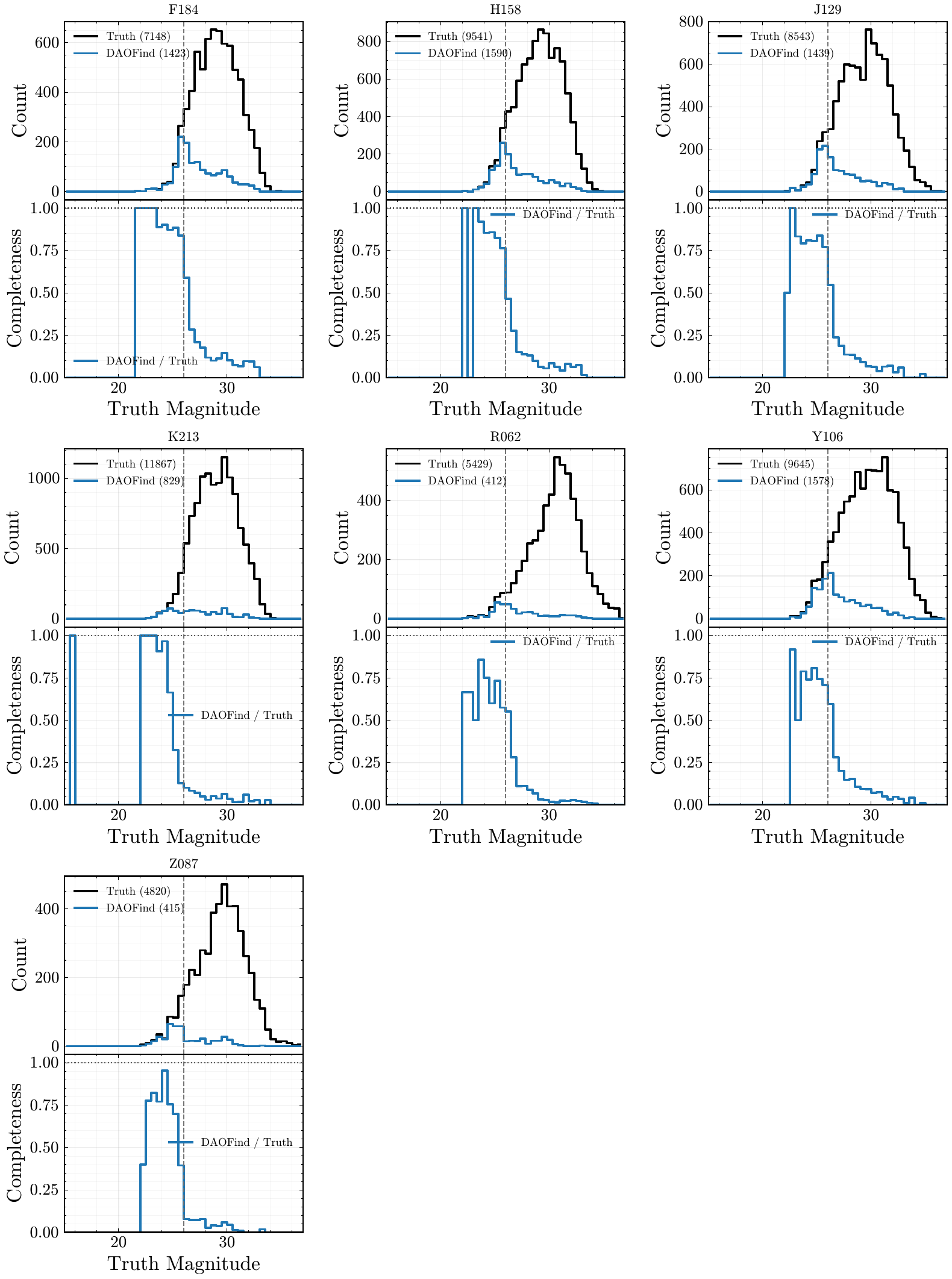}
    \caption{Filter-wise detections by the DAOFind algorithm in the OU24 dataset. The DAOFind numbers are all positive detections. Vertical lines are shown at a mag of 26. As described in the text, the models we present are currently based on this value as a limiting magnitude irrespective of filter and exposure, and our ground truth numbers are thus a fraction of the DAOFind numbers and are shown in Table~\ref{tab:dataset-filterwise-distribution}.}
    \label{fig:detections26}
\end{figure*}

We also included an additional set of features \feats in our model/dataset which will be used to train our model. These are the Magnitude Difference, Sharpness, two Roundness values, Peak value at the transient location, and are calculated by the photutils implementation of the DAOFind Algorithm on the \authz{ZOGY diff of the image}\footnote{Exact parameter names are: magnitude difference, roundness1, roundness2, sharpness and peak} (See Figure~\ref{fig:detections-eg}).

So our dataset $\mathcal{D}$ can be formally written as:

\begin{multline}    
    \mathcal{D} =\left\{ \left( x^{(i)}, \mathcal{F}^{(i)}, y^{(i)} \right) \ \middle|\right. \\ \left.\ x^{(i)} \in \mathbb{R}^{L \times L \times 4},\ \mathcal{F}^{(i)} \in \mathbb{R}^5,\ y^{(i)} \in \{0,1\} \right\}_{i=1}^N
\end{multline}

where $x^{(i)}$ is the cutout centered at the transient, \feats is the additional set of features as described above, $y^{(i)}$ is the ground truth label, and $N$ is the total number of detections.

Running these detection algorithms on the simulations with the default parameters, the Real to Bogus detections ratio is very low for both SExtractor and the DAOFind Algorithm. As partly shown in Figure~\ref{fig:detections-vs-mag}, there are a lot more bogus events that are detected. More precisely, the bogus-to-real ratio for the DAOFind Algorithm was close to 400 (381.69), and the bogus-to-real ratio for SExtractor was close to 200 (194.09). Moreover even though the DAOFind algorithm is detecting more transients than SExtractor, there are many spurious detections that happen to be in the vicinity of true transients \authz{by chance coincidence. To counter this, we regard all detections where the magnitude of the cross-matched source is fainter than 26, roughly corresponding to the 10$\sigma$ detection limit of the difference images, as false detections to avoid contamination of the true-positive sample of transients. The cut-off at 26 mag is not absolute, and will vary somewhat by exposure as well as by filters, and definitely for real data. But given that our methodology is largely filter-agnostic, we currently use a single simplifying number, and will incorporate filter- and exposure-wise numbers in future versions. See Figure~\ref{fig:detections26} to see filter-wise DAOFind detections compared to ground truth.} 

\section{Model Architecture}\label{sec:arch}

\begin{figure*}
    \centering
    \includegraphics[width=1\linewidth]{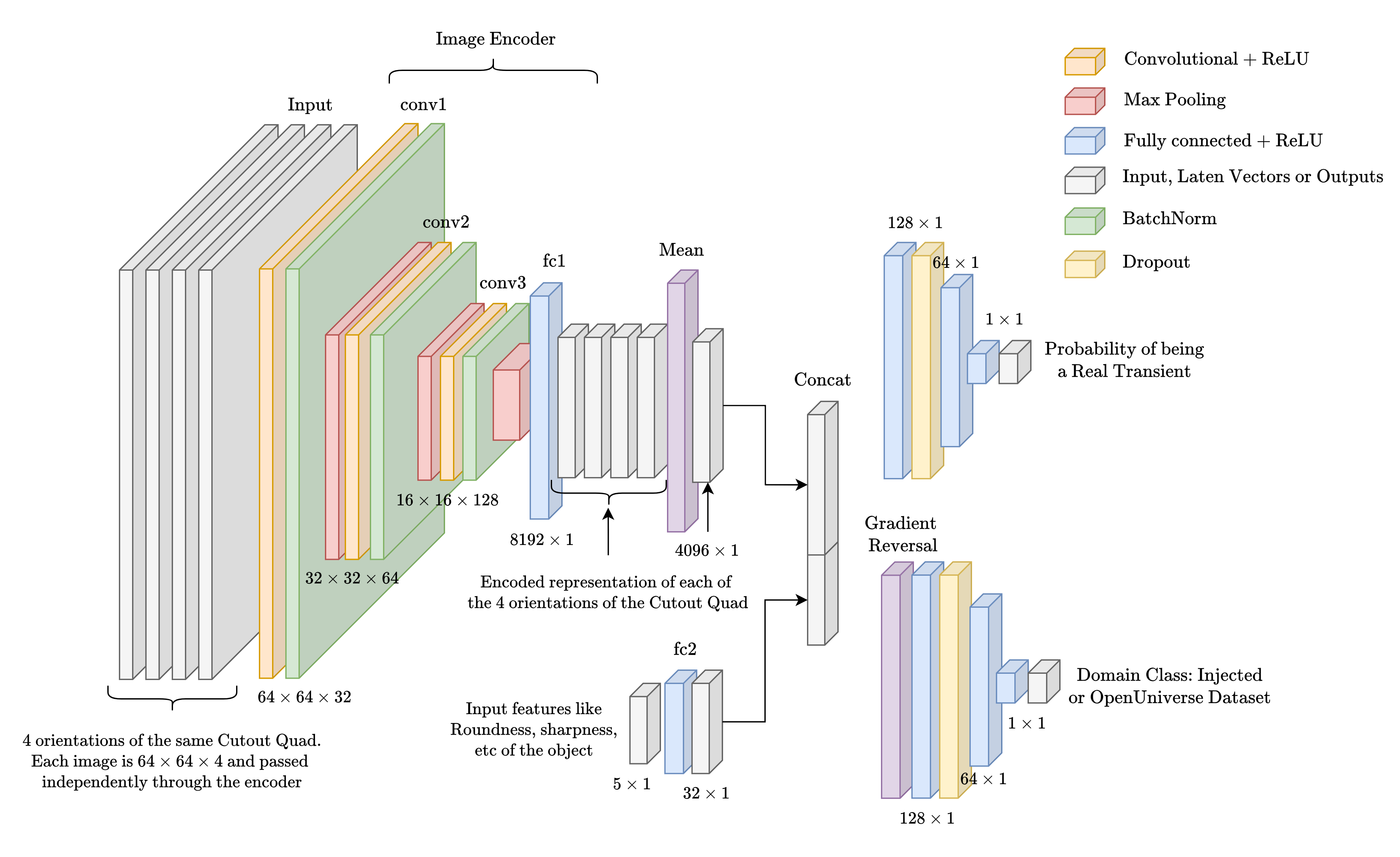}
    \caption{Model architecture for RuBR (Rotational Invariant CNN with Feature encoder). The intermediate dimensions of the vector are written next to each layer. \authz{The initial layers use ReLU as it is fast and does not suffer from vanishing gradients. The final layer uses a sigmoid activation, unlike the preceding layers, as it produces an interpretable, bounded score representing the probability that a given object is a real detection.}}
    \label{fig:model-arch}
\end{figure*}

The final real-bogus model consists of four parts (see Figure~\ref{fig:model-arch}): an image encoder, a feature encoder, a classifier, and a domain predictor (See Section~\ref{sec:domain-adv}). Since the objects (transients and variables) in the images captured by Roman will not have a preferred orientation, predictions should remain consistent under rotational transformations. To enforce this and to prevent the model from latching onto spurious directional features that may appear in the training dataset, we make the model's outputs invariant to the input orientation (see Appendix~\ref{sec:ml-back} for a short overview of the basic ML concepts used in the section). 

\subsection{The Image Encoder}
The image encoder is designed to produce rotation-invariant representations. Let 
\[
f: \mathbb{R}^{L \times L \times C} \rightarrow \mathbb{R}^{d_1}
\]
be a base encoder that maps an input image (or quad) to a feature embedding. Let 
\[
\mathrm{rot}: \mathbb{R}^{L \times L \times C} \times \mathbb{R} \rightarrow \mathbb{R}^{L \times L \times C}
\]
denote a rotation operator such that \(\mathrm{rot}(x, \theta)\) rotates the input image \(x\) by an angle \(\theta\).

We define a rotation-invariant encoder \(\tilde{f}\) as:
\begin{equation}
    \tilde{f}(x^{(i)}) = \frac{1}{360^\circ} \int_{0^\circ}^{360^\circ} f\big(\mathrm{rot}(x^{(i)}, \theta)\big)\, d\theta
\end{equation}

By construction, \(\tilde{f}\) satisfies rotational invariance:
\begin{equation}
    \tilde{f}(\mathrm{rot}(x^{(i)}, \phi)) = \tilde{f}(x^{(i)})
\end{equation}

In practice, we approximate this integral using a discrete set of rotations:
\begin{equation}
    \tilde{f}(x^{(i)}) = \frac{1}{N} \sum_{r=0}^{N-1} f\left(\mathrm{rot}\left(x^{(i)}, r \cdot \frac{360^\circ}{N}\right)\right)
\end{equation}

This approximation enforces invariance over \(N\) sampled orientations. In our implementation, we use \(N = 4\) (i.e., rotations at multiples of \(90^\circ\)) only to simplify the implementation and to avoid dealing with interpolation artifacts.

The base encoder $f$ is implemented as a three-layer convolutional neural network followed by a fully connected layer. Each convolutional block consists of a convolutional layer with ReLU activation, followed by batch normalization and max pooling.

\subsection{The Feature encoder}

In addition to image data, we incorporate additional features which are generated by the DAOFind algorithm \(\mathcal{F}^{(i)} \in \mathbb{R}^5\). These features are processed using a feature encoder:
\[
g: \mathbb{R}^5 \rightarrow \mathbb{R}^{d_2}
\]

The resulting feature embedding is:
\[
e^{(i)}_2 = g(\mathcal{F}^{(i)}), \quad e^{(i)}_2 \in \mathbb{R}^{d_2}
\]

The encoder \(g\) is a two-layer MLP with ReLU activations, mapping the input from dimension 5 to 16, and then to \(d_2 = 32\).




\subsection{The Classifier and the Domain Predictor}

The final representation is obtained by concatenating the image and feature embeddings:
\[
e^{(i)} = \mathrm{concat}\big(e^{(i)}_1,\, e^{(i)}_2) \in \mathbb{R}^{d_1 + d_2}
\]

This representation is passed through a classifier:
\[
c: \mathbb{R}^{d_1 + d_2} \rightarrow [0,1]
\]
which is implemented as a fully connected layer with a sigmoid activation. The predicted probability is:
\[
\hat{y}^{(i)} = c(e^{(i)})
\]

\authz{This is our \textit{RuBR} base architecture. We elaborate on the specific models we train and present in subsequent sections}. We also include a domain predictor used for domain-adversarial training. It shares the same architecture as the classifier but operates on features passed through a gradient reversal layer. Further details are provided in Section~\ref{sec:domain-adv}.

\section{Training Details}\label{sec:training}

The model was trained using a weighted binary cross-entropy loss to account for the inherent class imbalance between true and false detections.
\begin{multline}
    \mathcal{L}_{\text{WBCE}} = - \frac{1}{N} \sum_{i=1}^{N} \Big[ \, w_{1} \, y_i \, \log(p_i) \; + \\\; w_{0} \, (1 - y_i) \, \log(1 - p_i) \, \Big]
\end{multline}
\authz{where $w_0$ and $w_1$ are the class weights for the negative and positive classes, respectively.}

We used the Adam optimizer \citep{kingma2014adam} without weight decay, as preliminary experiments indicated no improvement in convergence with regularization. The initial learning rate was set to $10^{-3}$ and was adaptively reduced during training using the \texttt{ReduceLROnPlateau} scheduler with a reduction factor of $0.5$ and a patience of 5 epochs, based on the validation loss.

To prevent overfitting, early stopping was applied with a patience of 10 epochs, restoring the model weights corresponding to the best validation performance. The model was trained for a maximum of 70 epochs with a mini-batch size of 512. The model losses converged after around 40-50 epochs, after which early stopping was triggered.

We also experimented with various normalization techniques, which included zscale, standard normalization, etc. The best accuracy was achieved when the inputs were normalized filter-wise, meaning that for each filter, the cutouts were standardized independently by subtracting the cutouts' pixel mean and dividing by their standard deviation, resulting in zero mean and unit variance for each Roman filter. \authz{The model performance did not vary much from one filter to another, so, from a generalization perspective, we trained a single combined model. The filter-wise distribution from which we randomly pulled our training and test sets is summarized in Table~\ref{tab:dataset-filterwise-distribution}. Since this model was trained on the combined OU24 and local injections datasets, we call it \rubrc hereafter. Details of the training and test sets for this and other models described later are provided in Table~\ref{tab:dataset-training}. In each case, a validation set of $15\%$ was split out from the training set.} The models were trained on an NVIDIA A2 GPU. It takes roughly 4 hours to train the model for around 50 epochs.

\begin{deluxetable*}{lrcccrr}
    \authz{\tablecaption{The filter-wise distributions of transients in a subset of the OU24 dataset. It is from this set that our train-test sets have been generated. \label{tab:dataset-filterwise-distribution}}
    \tablehead{
      \colhead{Filter} &
      \colhead{Total transients} &
      \colhead{Ground Truth}
       & &
      \multicolumn{3}{c}{DAOFind Detections}\\
      \cline{3-3}
      \cline{5-7}\\[-12pt]
      \colhead{} &
       &
      \colhead{mag $\le$ 26} & &
      \colhead{True Positives} &
      \colhead{False Positives} &
      \colhead{Total}
    }
    \startdata
     F184 & 7,148  & 482 && 417   & 225,819   & 226,236   \\
     H158 & 9,541  & 725 && 588   & 214,442   & 215,030   \\
     J129 & 8,543  & 715 && 577   & 206,003   & 206,580   \\
     K213 & 11,867 & 742 && 294   & 180,768   & 181,062   \\
     R062 & 5,429  & 242 && 158   & 79,525    & 79,683    \\
     Y106 & 9,645  & 759 && 567   & 186,797   & 187,364   \\
     Z087 & 4,820  & 404 && 252   & 67,638    & 67,890    \\
     \hline
     \textbf{Total} & \textbf{56,993} & \textbf{4,069} & & \textbf{2,853} & \textbf{1,160,992} & \textbf{1,163,845} \\
    \enddata
    \tablecomments{Transients that are detectable ($\le 26$ Mag) form a part of our ground truth dataset of transients. As you can see, DAOFind produces many false positives, which we want to filter out.}}
\end{deluxetable*}

\begin{deluxetable}{lcr rr r}
   \authz{ \tablecaption{Sizes of training and testing datasets for \rubrc, \rubrl, \rubrd\label{tab:dataset-training}. From the training sets 15\% objects were carved out as validation sets.} 
    \tablehead{
      \colhead{Model} &
      \colhead{Type} &
      \multicolumn{2}{c}{Real} &
      \colhead{Bogus} &
      \colhead{Total} \\
      \cline{3-4} \\[-12pt]
      \colhead{} &
      \colhead{} &
      \colhead{OU24} &
      \colhead{Inj} &
      \colhead{} &
      \colhead{}
    }
    \startdata
     \rubrc & Train & 1222   & 9057   & 189721  & 200000  \\
     \rubrc & Test  & 3814   & 27111  & 569075  & 600000  \\
     \hline
     \rubrl & Train & --     & 9120   & 190880  & 200000  \\
     \rubrl & Test  & 177    & 3771   & 153064  & 157012  \\
     \hline
     \rubrd & Train & --     & 4560   & 95440   & 100000  \\
     \rubrd & DA    & 442    & --     & 99558   & 100000  \\
     \rubrd & Test  & 177    & 3771   & 153064  & 157012  \\
    \enddata
    }
\end{deluxetable}

\section{Results} \label{sec:exp-results}

\subsection{Evaluation Metrics}

\begin{figure}
    \centering
    \includegraphics[width=.8\linewidth]{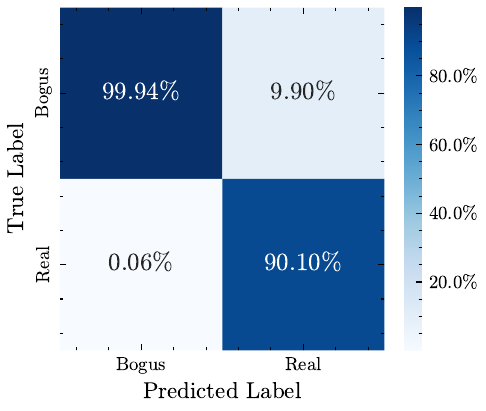}
    \caption{Normalized Confusion Matrix for the RuBR at a threshold of 0.58 (threshold chosen such that the F1 score is maximum on the validation dataset). Percentages correspond to the proportion of predictions belonging to the real or bogus class.}
    \label{fig:cm-at-0.575}
\end{figure}

Due to the significant class imbalance in the dataset, we evaluate model performance using the F1 score, which balances \textit{precision} and \textit{recall}. Accuracy can be misleading in such settings, as it may be dominated by the majority class, whereas the F1 score provides a more informative assessment by accounting for both false positives and false negatives. Precision is defined as the proportion of correctly predicted positive samples among all samples predicted as positive, given by
\[
\text{Precision} = \frac{TP}{TP + FP},
\]
where $TP$ and $FP$ denote true positives and false positives, respectively. Recall, on the other hand, measures the proportion of correctly predicted positive samples among all actual positive samples, defined as 
\[
\text{Recall} = \frac{TP}{TP + FN},
\]
where $FN$ represents false negatives. The F1 score combines both precision and recall as their harmonic mean:
\[
\text{F1 Score} = 2 \times \frac{\text{Precision} \times \text{Recall}}{\text{Precision} + \text{Recall}}.
\]
\authz{A concise summary of the fractions of real and bogus transients identified by \rubrc is presented as a confusion matrix in Figure~\ref{fig:cm-at-0.575}}.

\begin{table*}
    \centering
    \caption{Performance comparison of different models for real bogus classification. \authz{The RuBR model here is the \rubrc since it was trained on the combined OU24 and local injections datasets.}}
    \begin{tabularx}{\textwidth}{lXXXX}
         \hline
         \hline
         Model & Precision & Recall & F1-score & Accuracy \\
         \hline
         DenseNet121 & 80.42\% & 51.04\% & 0.6244 & 99.85\%\\
         VGG11 & 73.78\% & 61.05\% & 0.6682 & 99.86\%\\
         ResNet-18 & 79.93\% & 67.06\% & 0.7293 & 99.88\%\\
         Deep-HITS & 85.91\% & 73.72\% & 0.7935 & 99.92\%\\
         Braai  & 90.23\% & 51.71\% & 0.6575 & 99.87\%\\
         RuBR (Ours) & \textbf{90.10\%} & \textbf{71.68\%} & \textbf{0.7984} & \textbf{99.92\%}\\
         \hline
    \end{tabularx}
    \raggedright
    \tablecomments{Deep-HITS and Braai are previously proposed models for real--bogus classification (see \citealt{cabrera2017deep,duev2019real}).}
    \label{tab:model-comparison-results}
\end{table*}

\subsection{Comparison with other widely used models}

We compare \rubrc against several widely used architectures, including CNN-based models such as DenseNet121, VGG11, and ResNet-18.
In addition, we benchmark our approach against established models specifically designed for Real–Bogus classification, such as Deep-HITS \citep{cabrera2017deep} and Braai \citep{duev2019real}. \authz{See Table~\ref{tab:model-comparison-results} for details}.

For a fair comparison, all models are trained for a maximum of 70 epochs, with an early stopping condition and a patience of 10 epochs. The loss for all models converged before 70 epochs. The dataset used for training is the same as described above. We apply filter-wise standard normalization to the dataset for all models to ensure consistency during training.

\subsection{Magnitude distribution of the Detected Transients}
\begin{figure}
    \centering
    \includegraphics[width=1\linewidth]{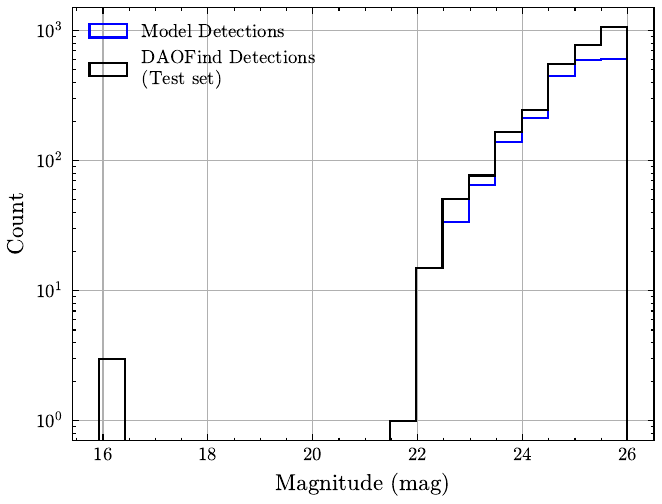}
    \caption{AB Magnitude Distribution of the cross-matched sources for the transients detected by \authz{\rubrc} and all the transients in the testing dataset.}
    \label{fig:magnitude_distribution}
\end{figure}
We analyze the magnitude distribution of all detected transients to assess the detection characteristics of the pipeline. Figure~\ref{fig:magnitude_distribution} shows the distribution of the ground-truth AB magnitudes corresponding to the cross-matched sources of each detected transient, along with the total number of transients identified by the DAOFind algorithm. This comparison helps in understanding the sensitivity of the detection pipeline across different brightness levels and in identifying any biases toward brighter or fainter sources. We observe that most of the relatively bright transients are detected, and the number of detections gradually decreases as we approach the detection limit of approximately $26^{\mathrm{th}}$ magnitude.

\subsection{Different thresholds for detecting transients}

\begin{table}
    \centering
    \caption{Precision, Recall, F1-score, and Accuracy of \authz{\rubrc} at different threshold values.}
    \begin{tabular}{ccccc}
         \hline
         \hline
         Threshold&  Precision&  Recall&  F1-score & Accuracy\\
         \hline
         0.3& 84.31\% & 74.43\% & 0.7906 & 99.92\%\\
         0.4& 87.48\% & 72.87\% & 0.7951 & 99.92\%\\
         0.5& 89.77\% & 71.82\% & 0.7980 & 99.92\%\\
         0.6& 91.89\% & 70.02\% & 0.7948 & 99.92\%\\
         0.7& 93.91\% & 67.50\% & 0.7855 & 99.92\%\\
         0.8& 96.22\% & 63.97\% & 0.7685 & 99.92\%\\
         0.9& 98.34\% & 58.47\% & 0.7334 & 99.91\%\\
         \hline
    \end{tabular}
    \label{tab:threshold-vs-metric}
\end{table}

\begin{figure}
    \centering
    \includegraphics[width=1\linewidth]{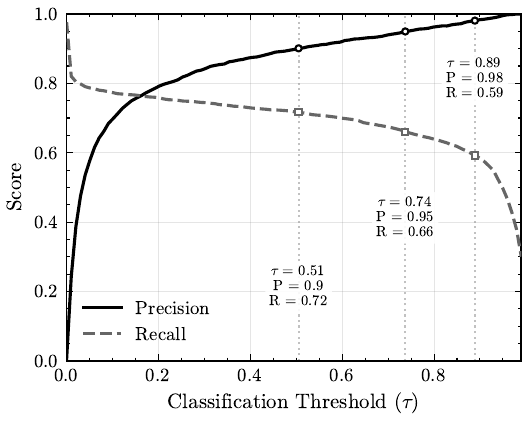}
    \caption{Precision-Recall curve vs threshold for \rubrc. A threshold of 0.58 gives us the best F1 score on the validation dataset. \authz{Increasing the threshold makes the model more precise, but reduces the number of detected transients. End-users can choose their own threshold using this curve on the rb-score for intended science.}}
    \label{fig:prec-recall-curve}
\end{figure}

\authz{\rubrc} outputs a continuous score between 0 and 1, \authz{called the rb-score,} representing the probability of a candidate being \authz{astrophysically real. A score closer to 1 indicates higher confidence that the detection is real, while values near 0 suggest it is likely bogus.} To convert these scores into binary classifications, we evaluated the model’s performance across different chosen decision thresholds, as summarized in Table~\ref{tab:threshold-vs-metric} \authz{and the overall trend between threshold and performance is illustrated in Figure~\ref{fig:prec-recall-curve}}. As the threshold increases, the model becomes more conservative in labeling detections as real, leading to higher precision but a corresponding decrease in recall. This trade-off reflects the balance between minimizing false positives and maintaining completeness in transient recovery. The F1-score remains relatively stable around 0.78 for thresholds between 0.4 and 0.7. The rb-score will be part of RAPID's alert packets, thereby allowing end-users to choose their own thresholds for intended science (for instance, purer samples, versus greater completeness).

\section{Adapting to real-world data} \label{sec:domain-adv}

\begin{deluxetable*}{p{0.30\textwidth}>{\centering\arraybackslash}p{0.15\textwidth}>{\centering\arraybackslash}p{0.15\textwidth}>{\centering\arraybackslash}p{0.15\textwidth}>{\centering\arraybackslash}p{0.15\textwidth}}
\tablewidth{\textwidth}
\tablecaption{Performance of the models with and without domain adaptation on the different datasets.
\label{tab:dann-performance}}
\tablehead{
  \multicolumn{1}{c}{} &
  \multicolumn{2}{c}{Injected Dataset} &
  \multicolumn{2}{c}{OpenUniverse2024 Dataset} \\
  \multicolumn{1}{c}{} &
  \multicolumn{2}{c}{} &
  \multicolumn{2}{c}{(without ground truth labels)} \\
  \cline{2-3} \cline{4-5}
  \colhead{} &
  \colhead{\% Precision} &
  \colhead{\% Recall} &
  \colhead{\% Precision} &
  \colhead{\% Recall}
}
\startdata
No Domain Adaptation \authz{(\rubrl)}  & 91 (91) & 56 (56)   & 35 (36)    & 54 (54)   \\
With Domain Adaptation \authz{(\rubrd)} & 84 (89) & 59 (60)   & 27 (49) & 85 (72) \\
\enddata
\tablecomments{\authz{Both the models were trained only on the ground truth labels of the injected dataset and tested not just on the injected dataset, but also on the OU24 simulated images, which produce different subtraction artifacts compared to the injected dataset, thereby better reflecting real-world conditions. We have reported the performance on both the test sets. Numbers are provided at a fixed threshold of 0.5, and, in brackets, at thresholds optimized to maximize F1.}}
\end{deluxetable*}

\begin{figure}
    \centering
    \includegraphics[width=1\linewidth]
    {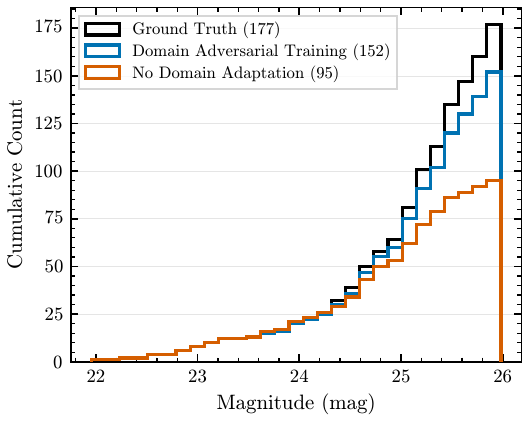}
    \caption{Cumulative histogram of the magnitude distribution of the transients detected with and without domain adaptation \authz{at a rb-score threshold of 0.5. While at brighter magnitudes both methods do equally well, identifying most of the transients, at fainter magnitudes, a model trained using domain-adversarial learning identifies a greater number of transients. The total number of transients detected by each method is indicated in parentheses, along with the total number of ground truth examples.}}
    \label{fig:dann-mag-hist}
\end{figure}


Once the \authz{Roman} telescope is launched, the real images are likely to exhibit artifacts that differ from those present in OU24 simulations. As a result, \authz{\rubrc} will need to be adapted to this shift in data distribution. Domain adversarial training provides a framework for such adaptation by enabling the model to learn domain-invariant representations without requiring ground truth labels for the target (real) data \citep[see, e.g.,][]{ganin2016domain}. This will be helpful in the initial stages of deployment, where we don't have ground truth labels and still will be able to provide reliable detections. Once we have ground truth labels for some of the objects that are identified, we can train the models to further improve accuracy.

\begin{figure}
    \centering
    \includegraphics[width=1\linewidth]
    {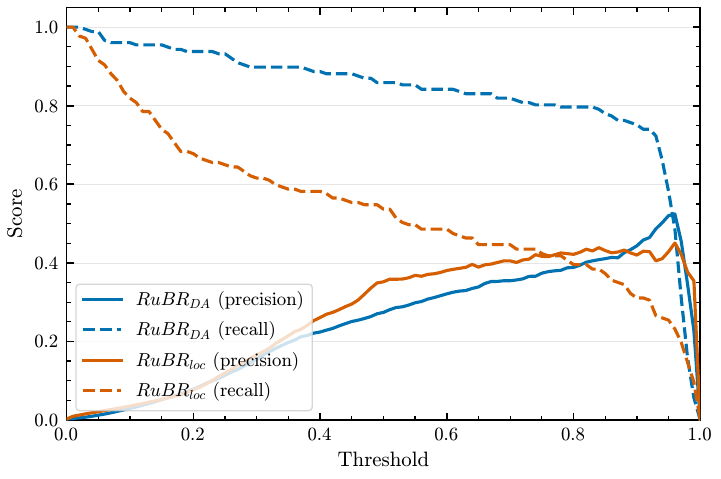}
    \caption{\authz{Precision-Recall curve vs threshold for
    \rubrl and \rubrd. As usual, changing the threshold will improve precision or recall at the cost of the other.}}
    \label{fig:prt_da}
\end{figure}

In our setup, we use the concatenated embeddings from the image encoder and the feature encoder, denoted by $e^{(i)}$ (see Figure~\ref{fig:model-arch}), as the shared representation learned by the network. On top of this representation, we add a domain classification head whose task is to predict whether the embedding comes from simulated data or from real observations. Training follows an adversarial procedure: the domain classifier is trained to distinguish between simulated and real embeddings, while the image and feature encoders are trained, through the gradient reversal layer, to make this distinction difficult. This encourages the encoders to learn representations that are similar across both the data (real and simulated), while still preserving the information needed for accurate real--bogus classification.

More formally, any domain adversarial training network consists of three components: a feature extractor \( G_f \), which in our case will be the final concatenated output of both the feature encoder and the image encoder, a label predictor \( G_y \), which is the classification head, and a domain classifier \( G_d \). For an input object with image \( x \), features \feats, label \( y \), and domain label \( d \in \{0,1\} \) (simulated or real), the overall objective is formulated as a minimax optimization problem:
\[
\min_{G_f, G_y} \max_{G_d} \ \mathcal{L}_y(G_y(G_f(x)), y) - \lambda \mathcal{L}_d(G_d(G_f(x)), d),
\]
where \( \mathcal{L}_y \) is the real-bogus loss (the cross-entropy for real-bogus classification) and \( \mathcal{L}_d \) is the domain classification loss. The gradient reversal layer (GRL) inverts the gradient from \( \mathcal{L}_d \) during backpropagation, encouraging the feature extractor \( G_f \) to learn representations that are discriminative for the main task yet indistinguishable across domains.

This adversarial mechanism ensures that the learned features are invariant to domain-specific differences between simulated and real images, allowing the model to retain its performance after deployment with minimal fine-tuning. Such an approach will be crucial for adapting Roman’s transient detection models once real observational data becomes available.

To test this, we took two different datasets of Roman simulated images. One consisting only of images with injected transients, and one consisting of only the images with transients, which were a part of the OU24 simulation (as a substitute for the real data). We trained two models, one without domain adaptation, simply training on the injected dataset as a control for the experiment (\authz{called \rubrl hereafter}), and one with domain adaptation, trained on the injected dataset and domain adapted to the OU24 dataset (\authz{called \rubrd hereafter}). \authz{\rubrl performs poorly on the OU24 dataset, giving a precision of 36\% and a recall of 54\%. However, when we trained the model with domain adaptation \authz{(\rubrd)}, an improvement is observed, increasing precision to 49\% and recall to 72\%} \authz{(See Table~\ref{tab:dann-performance})}. \authz{Moreover, we observe that applying domain adaptation enhances the model’s ability to detect fainter transients (see Figure~\ref{fig:dann-mag-hist}), retaining a high performance at the brighter end. As with \rubrc, users can choose their own threshold for higher precision or recall (See Figure~\ref{fig:prt_da}).}\authz{The final accuracy of the domain classification was close to 50\%, showing that the feature encoder was successfully adapted to the OU24 dataset.}

Once we have a robust training dataset for the real Roman data, made using human-assisted labels (citizen science, etc.), these numbers will improve. However, till then, this method gives us a promising approach for both generating a good training dataset as well as providing reliable detections in the early stages of Roman observations.

\section{Ablation studies}\label{sec:ablation}

\begin{deluxetable}{lcccc}
    \tablecaption{Performance of different ablations of the model at a threshold of 0.5.\label{tab:ablation-tab}}
    \tablehead{
      \colhead{Modification} &
      \colhead{Precision} &
      \colhead{Recall} &
      \colhead{F1-score}
    }
    \startdata
    No Modifications & \textbf{90.10\%}  & 71.68\% & \textbf{0.7984} \\
    No Rotational Invariance & 51.68\% & 78.03\% & 0.6218 \\
    Standard Normalisation & 78.58\% & 70.87\% & 0.7452 \\
    No Feature Encoder & 77.55\% & \textbf{78.34\%}  & 0.7794 \\
    \enddata
    \tablecomments{Removing rotational invariance, filter-wise normalization, or the feature encoder leads to degraded performance.}
\end{deluxetable}

\subsection{Effectiveness of Rotational Invariance}

Since the output of the model should remain consistent irrespective of the image orientation, we introduced rotational invariance into the architecture. This modification ensures that the model’s predictions are robust to rotations of the input images, aligning with the physical symmetry of astronomical sources. Incorporating rotational invariance led to a notable improvement in the model’s performance. The quantitative impact of this addition is summarised in Table~\ref{tab:ablation-tab}, which compares the results of the model trained with and without rotational invariance.

\subsection{Effectiveness of the Feature encoder}

To evaluate the contribution of the feature encoder, we compared the transients detected by our model with and without the feature encoder. We trained two models, one with the feature encoder and one without it, on the same dataset. The two models share a substantial fraction of detections, approximately 53.8\% of all the detected transients (by at least one model), indicating that both capture a consistent subset of genuine transients. However, the model with the feature encoder identifies an additional 46.2\% of detections, in addition to those detected by the image encoder model alone, highlighting the complementary information captured by the learned feature representations. This demonstrates that incorporating additional features computed by DAOFind, along with image-based features, improves the model’s ability to generalize and identify transients that may be missed by a purely rotationally invariant approach.

\subsection{Effectiveness of doing filter-wise normalization}

The motivation for applying filter-wise normalization arises from the fact that images obtained through different filters exhibit distinct intensity distributions, as the flux received from astronomical sources varies across wavelength bands. Consequently, each filter’s images possess different mean and variance characteristics. To account for these differences, we normalize the data separately for each filter rather than applying a global normalization across all bands. Empirically, we observed that models trained on filter-wise normalized data achieved improved performance compared to those trained on data normalized jointly across filters (See Table \ref{tab:ablation-tab}), indicating that preserving the relative photometric characteristics of each band enhances the model’s ability to learn meaningful representations.

\section{Conclusion and Future Work} \label{sec:future-work}

In this work, we developed and validated a rotationally invariant convolutional neural network that integrates both image and feature encoders to distinguish genuine astrophysical transients from spurious detections. By enforcing rotational invariance through orientation-averaged embeddings, the proposed architecture achieves robustness to arbitrary image rotations, a critical requirement for astronomical imaging. When trained and evaluated on the OU24 simulations, the model attains a precision of 90.10\% and a recall of 71.68\% \authz{(Table~\ref{tab:model-comparison-results})}.

The comparatively lower recall can be attributed to multiple factors. First, the point spread function (PSF) model used during source extraction is not optimal for the reference images, which can lead to imperfect flux estimation and subtraction artifacts. Second, the injected transients exhibit significant contamination from residuals near galaxy cores, where background structures and subtraction artifacts are prevalent. As a result, a non-negligible fraction of true transients are difficult to distinguish from subtraction residuals. A promising direction for future work is the construction of a cleaner training set by excluding injected transients located within a specified radial threshold from galaxy centers, thereby reducing confusion from host-galaxy residuals and improving recall.

Beyond its current use with simulated data, we propose adopting a domain-adversarial fine-tuning framework to bridge the gap between simulated and real Roman Space Telescope data once available. This approach would enable the model to adapt to real detector characteristics and observational systematics while preserving its discriminative capability, ensuring reliable performance under operational conditions.

Several additional limitations and avenues for improvement remain. The current OU24 dataset contains only transient events and does not include variable sources, such as periodic or semi-periodic objects. Incorporating variable objects would provide a more realistic representation of the sky and enable a more rigorous evaluation of the model’s ability to distinguish true transients from intrinsic variability. Furthermore, additional optimization of source-extraction parameters (e.g., in \texttt{DAOFind} or related pipelines) is likely to yield performance gains, as the present dataset is highly imbalanced, with a large number of bogus detections relative to true events. Addressing these limitations will further strengthen the applicability of the proposed framework for automated transient filtering in Roman’s alert pipeline.

The Roman products that RAPID generates will be public, so are the OU24 data. In line with that, to support open science and reproducibility, we are making our trained models publicly available\footnote{Code can be found at 
\url{https://github.com/Caltech-IPAC/rapid/tree/main/RuBR}
}. This will allow other researchers to experiment with the models and explore new ideas, while we continue to improve them in parallel. The model output confidence scores will be included in the RAPID alert stream, so that astronomers can better judge the reliability of each alert when deciding on follow-up observations.

\begin{acknowledgments}
RAPID project infrastructure team acknowledges NASA support under awards 80NSSC24M0020 and 80NSSC26K0019 (program NNH22ZDA001N-ROMAN). Use of AI models was made in a small number of places as follows: ChatGPT-5.5 (chatgpt.com) for rephrasing a few sentences for clarity and for formatting LaTeX tables, Claude 4.6 Sonnet (claude.ai) for formatting tables and fixing LaTeX errors, and ChatGPT 5.2 Codex (Copilot agent mode) for moving the necessary files to a separate repository and refactoring them for better readability. We thank Aayush Kuloor for verifying the \textit{RuBR} models' results.
\end{acknowledgments}

\begin{contribution}
KG conceived and developed the machine-learning approach, designed the model architecture, carried out the experiments, proposed the domain-adaptation methodology, analyzed the results, and wrote the first draft of the manuscript. AM mentored and supervised the project, helped shape the research direction and methodology with the big picture in mind, contributed to the interpretation of results, and played a major role in revising and improving the manuscript. JJ co-mentored the project and provided guidance on image differencing, data preparation, and the nitty-gritty of injected data. RRL developed and maintained the RAPID image-differencing pipeline used in this work. BR, LY, RML, SVD, and MK contributed to discussions and provided feedback as members of the RAPID PIT. 
\end{contribution}

\facilities{Nancy Grace Roman Space Telescope}

\software{tensorflow \citep{tensorflow2015-whitepaper},
          pytorch \citep{paszke2019pytorch},
          astropy \citep{2013A&A...558A..33A,2018AJ....156..123A,2022ApJ...935..167A},  
          Source Extractor \citep{1996A&AS..117..393B},
          photutils \citep{larry_bradley_2025_17129028},
          numpy \citep{harris2020array},
          pandas \citep{reback2020pandas},
          sklearn \citep{scikit-learn}
          }

\appendix

\section{Machine Learning Background}\label{sec:ml-back}

Machine learning aims to learn patterns from data in order to make predictions on new, unseen examples \citep{bishop2006pattern}. A typical setup consists of three main components: (1) a data set containing input examples and, when available, corresponding labels; (2) a model, which is a parameterized mathematical function; and (3) a training procedure that adjusts the model parameters so that the model's predictions better match the desired outputs. During training, the model produces predictions that are compared with the true labels using a loss function. An optimization algorithm such as stochastic gradient descent (SGD; \citealt{lecun1998sgd}) or Adam \citep{kingma2014adam} is then used to update the parameters in the direction that reduces this loss. Once trained, the model can be applied to new data for tasks such as classification, regression, or detection.

\subsection{Artificial Neural Networks}

Artificial neural networks (ANNs) \citep{rumelhart1986learning} are composed of basic computational units called neurons. Each neuron receives an input vector $\mathbf{x}$, computes a weighted sum using a weight vector $\mathbf{w}$ and a bias term $b$, and applies a non-linear activation function~$f$:
\begin{equation}
    y = f(\mathbf{w}^\top \mathbf{x} + b)
\end{equation}
Neurons are arranged in layers, and stacking multiple such layers yields a multilayer perceptron (MLP). MLPs can approximate complex mappings between inputs and outputs \citep{hornik1989multilayer}, although they do not explicitly exploit spatial structure in image data.

\subsection{Convolutional Neural Networks}

Convolutional neural networks (CNNs) extend ANNs to data with spatial structure and have become the standard architecture for image analysis tasks \citep{lecun1998mnist}. CNNs contain several specialized components described below.

\subsubsection{Convolution Layers}

A convolution layer consists of small parameter matrices called \emph{filters} or \emph{kernels}. 
Given an input array~$x$, each filter~$W$ slides across the spatial dimensions and computes a weighted sum over local patches:
\begin{equation}
    (W * x)(i,j) = \sum_{u,v} W(u,v)\, x(i+u, j+v)
\end{equation}
producing a feature map that highlights locations where the learned pattern appears. 
Multiple filters are applied simultaneously, allowing the layer to detect a variety of local features \citep{lecun1998mnist}.

\subsubsection{Max-Pooling}

Pooling layers downsample feature maps by selecting a representative value within non-overlapping windows. 
In \emph{max-pooling}, the output at each window is the maximum value:
\begin{equation}
    y_{i,j} = \max_{(u,v) \in \mathcal{W}_{ij}} x_{u,v}
\end{equation}
where $\mathcal{W}_{ij}$ denotes the window associated with output position $(i,j)$. 
Pooling reduces spatial resolution, improves computational efficiency, and provides a degree of translation invariance \citep{boureau2010theoretical}.

\subsubsection{Batch Normalization}

Batch normalization \citep{ioffe2015batch} stabilizes and accelerates training by normalizing intermediate activations within each mini-batch and then allowing the network to learn an appropriate scale and shift. 
For a mini-batch of activations $\{x_1,\dots,x_m\}$ (for a single feature/channel), compute the batch mean and variance:
\begin{align}
    \mu_{\text{B}} &= \frac{1}{m}\sum_{i=1}^m x_i, \\
    \sigma_{\text{B}}^2 &= \frac{1}{m}\sum_{i=1}^m (x_i - \mu_{\text{B}})^2.
\end{align}
The activations are normalized with a small constant $\varepsilon>0$ for numerical stability:
\begin{equation}
    \hat{x}_i = \frac{x_i - \mu_{\text{B}}}{\sqrt{\sigma_{\text{B}}^2 + \varepsilon}}.
\end{equation}
Two \emph{learnable} parameters per channel, $\gamma$ (scale) and $\beta$ (shift), are then applied:
\begin{equation}
    y_i = \gamma\,\hat{x}_i + \beta.
\end{equation}
During training, running estimates of the mean and variance are maintained to be used at inference time. These running statistics are updated by exponential moving averages. At inference time the batch statistics are replaced by these running statistics so that the layer performs a fixed affine transform of its inputs.

\subsubsection{Activation Functions: ReLU}

Activation functions introduce non-linearity into neural networks. The rectified linear unit (ReLU) \citep{nair2010relu} is defined as $f(x) = \max(0, x)$ and is widely used due to its computational efficiency and favorable training dynamics compared to classical activations such as sigmoids or hyperbolic tangents.

\subsubsection{Multilayer Perceptron (MLP)}

At the final stage of many CNN architectures, the learned features are flattened and passed into an MLP composed of fully connected layers. In these layers, every neuron receives input from all outputs of the previous layer, allowing the network to combine and interpret the extracted features to produce a final prediction.

Taken together, we can abstract out a combination of the above-mentioned components as a single parametric function
\[
f_\theta : \mathcal{X} \rightarrow \mathcal{Y},
\]
where $\mathcal{X}$ denotes the input space (e.g.\ image cutouts), $\mathcal{Y}$ the output space (e.g.\ class labels or scores), and $\theta$ the collection of all learnable parameters of the network, including convolutional filters, normalization parameters, and weights of fully connected layers. Training the model corresponds to finding parameter values $\theta$ that minimize a chosen loss function over the training data, such that the network produces outputs consistent with the desired predictions. Once optimized, the same function can be applied deterministically to unseen inputs to perform inference.

\bibliography{main}{}
\bibliographystyle{aasjournalv7}

\end{document}